\documentclass[10pt, conference, compsocconf]{IEEEtran}

\usepackage{algorithm}
\usepackage{algpseudocode}
\usepackage{amsmath}
\usepackage{amssymb}
\usepackage{booktabs}
\usepackage{hyperref}
\usepackage{ifthen}
\usepackage{keyval}
\usepackage{mathtools}
\usepackage{placeins}
\usepackage{siunitx}
\usepackage{tabu}
\usepackage{tikz}
\usepackage{upgreek}

\usetikzlibrary{arrows.meta, backgrounds, calc, fit, positioning}

\DeclareMathOperator*{\binop}{\oplus}

\begin{document}

\newcommand{\thetavec}{\ensuremath{\boldsymbol{\theta}}}
\newcommand{\fusop}{\ensuremath{\mathcal{F_{\thetavec}}}}
\newcommand{\mutanfeat}[1]{\ensuremath{\tilde{\mathbf{#1}}}}
\renewcommand{\m}{\ensuremath{\mathbf{m}}}
\newcommand{\n}{\ensuremath{\mathbf{n}}}
\newcommand{\q}{\ensuremath{\mathbf{q}}}
\renewcommand{\v}{\ensuremath{\mathbf{v}}}
\newcommand{\z}{\ensuremath{\mathbf{z}}}
\renewcommand{\A}{\ensuremath{A}}
\newcommand{\B}{\ensuremath{B}}
\newcommand{\C}{\ensuremath{C}}
\newcommand{\R}{\ensuremath{\mathbb{R}}}
\newcommand{\T}{\ensuremath{\mathcal{T}}}
\newcommand{\Tfibre}{\ensuremath{\T_{i_1 \cdots i_{n - 1} \,:\, i_{n + 1} \cdots i_N}}}
\newcommand{\TsizeIn}[1]{\ensuremath{I_1 \times \cdots \times I_{n - 1} \times #1 \times I_{n + 1} \times \cdots \times I_N}}
\newcommand{\binopb}{\ensuremath{\sideset{}{_b}\binop}}
\newcommand{\binopseq}{\ensuremath{{(\binopb)}_{b = 1}^B}}
\newcommand{\binoppartition}{\ensuremath{{\{\tuckbranch\}}_b}}
\newcommand{\binoppartB}{\ensuremath{\mathbb{B}}}
\newcommand{\binoppartBseq}{\ensuremath{{(\binoppartB_b)}_{b = 1}^B}}
\newcommand{\tuckbranch}{\ensuremath{\T_r^{\q{}\v{}}}}
\newcommand{\hadamardqv}{\ensuremath{N_r\mutanfeat{q} \odot M_r\mutanfeat{v}}}

%
\title{Generalized Hadamard-Product Fusion Operators\\
for Visual Question Answering}


\author{
\IEEEauthorblockN{Brendan Duke\IEEEauthorrefmark{1}\IEEEauthorrefmark{2}, Graham W.~Taylor\IEEEauthorrefmark{1}\IEEEauthorrefmark{2}\IEEEauthorrefmark{3}}

\IEEEauthorblockA{\IEEEauthorrefmark{1}School of Engineering,
University of Guelph\\
}
\IEEEauthorblockA{\IEEEauthorrefmark{2}Vector Institute for Artificial Intelligence
}
\IEEEauthorblockA{\IEEEauthorrefmark{3}Canadian Institute for Advanced Research
}
\{bduke,gwtaylor\}@uoguelph.ca
}
\maketitle

\begin{abstract}
We propose a generalized class of multimodal fusion operators for the task of
visual question answering (VQA). We identify generalizations of existing
multimodal fusion operators based on the Hadamard product, and show
that specific non-trivial instantiations of this generalized fusion
operator exhibit superior performance in terms of OpenEnded accuracy on
the VQA task. In particular, we introduce Nonlinearity Ensembling,
Feature Gating, and post-fusion neural network layers as fusion
operator components, culminating in an absolute percentage point
improvement of~$1.1\%$ on the VQA~2.0 test-dev set over baseline fusion
operators, which use the same features as input. We use our findings as
evidence that our generalized class of fusion operators could lead to
the discovery of even superior task-specific operators when used as a
search space in an architecture search over fusion operators.
\end{abstract}

\begin{IEEEkeywords}
Model Selection; Visual Question-Answering
\end{IEEEkeywords}

%
\IEEEpeerreviewmaketitle{}

\section{Introduction}

Multimodal applications offer a challenge to model selection in machine
learning, as the interactions between different data modalities (e.g.,~between
video and audio, or between images and questions) may require a complicated
prior in order to accurately capture regularities necessary for downstream tasks.

The particular multimodal application that we consider in this work is that of
visual question-answering~(VQA), i.e., of producing a natural language response
to the combined input consisting of an image and a natural language question
pertaining specifically to that image. In the case of VQA, the complexity of
the task exists in both extracting useful feature representations from the
question and the image, as well as the ``fusion'' of these feature
representations by combining them in order to predict the answer to the
question. In this paper, we focus on the problem of combining feature
representations in the VQA task through models based on a generic ``fusion
operator'' definition.

We illustrate the complexity of model selection for the VQA task by designing a
class of multimodal fusion operators, each of which combines raw question and
visual data streams to predict answers to the given questions based on an
image. We evaluate specific instances of high performing fusion operators
belonging to the same design class.

We evaluate and discuss three multimodal fusion architectural components that
emerge as improving performance as part of the investigation into the general
class of fusion operators:

\begin{enumerate}
        \item The use of a gating mechanism, wherein individual features
                extracted by the fusion operator are turned on or off by a
                multiplicative interaction.

        \item The introduction of distinct nonlinearities between parallel
                components in the fusion operator, which we hypothesize adds a
                performance boost due to an ensembling effect.

        \item The additional introduction of learned nonlinearity in the form
                of a neural network inside the fusion operator, which takes
                features from the bilinear interaction of a pair of question
                and visual feature vectors as input.
\end{enumerate}

\section{Related Work}

\subsection{Model Selection}

We propose that the task of model selection applied to multimodal
problem domains can be improved by using automated architecture search techniques.
Reinforcement learning has been used to conduct automated search for neural
network architectures~\cite{zoph2016neural}, gradient
descent optimizers~\cite{bello2017neural} and activation
functions~\cite{ramachandran2017searching}. Other options for architecture
search include evolutionary optimization~\cite{liu2018hierarchical}, Bayesian
optimization~\cite{malkomes2016bayesian}, and gradient descent-based methods.

We contribute to multimodal model selection by describing a generalized class
of fusion operators that can be used as a design space for many of the
search techniques described above. We enable future research to bypass
the difficult problem of model selection for multimodal applications by using
automated model design techniques that use the search space we describe as a
basis.

\subsection{Fusion Operators}

We focus on multimodal fusion of two modalities, applied to the task of visual
question-answering. We use modality to refer to a raw data stream of
information, as would be presented to an observer from a sensor. In the case of
the VQA application, the first modality is the sentence corresponding to the
asked question. The second modality is the image about which the question is
being asked.

Current research in multimodal fusion for VQA focuses
on performance improvements by approximating a bilinear product between a
feature vector~$\q$ extracted from the question, and a feature vector~$\v$
extracted from the
image~\cite{DBLP:conf/emnlp/FukuiPYRDR16, Kim2017, ben2017mutan}.  The feature
vectors $\q$ and $\v$ can be produced by pre-trained feature extraction methods
specific to encoding information from the sentence and image modalities into
vector representations.

In the experimental design of this work, as well as in the work
of~\cite{Kim2017} and~\cite{ben2017mutan}, a pre-trained Residual Network
model~\cite{he2016deep} is used as a feature extractor for the image data
stream, and a pre-trained Skip-Thought
Vectors~\cite{DBLP:journals/corr/KirosZSZTUF15} model is used to extract
features from the sentence data stream.

In general, we define a fusion operator for the VQA task as a function
$\fusop$, parametrized by~$\thetavec$, of the question feature vector~$\q$ and
the visual feature vector~$\v$. The fusion operator~$\fusop$ computes a vector
output, which is consumed downstream by a function~$g$, e.g.\ a linear layer
followed by a softmax layer, in order to model a probability distribution over
the answer~$y$ conditional on~$\q$ and~$\v$:

\begin{equation}
        p(y \mid \q \,, \v \,; \thetavec) = g(\fusop{(\q \,, \v)}) \,.\
\label{eqn:general-fusion-op}
\end{equation}

The methods of~\cite{DBLP:conf/emnlp/FukuiPYRDR16},
\cite{Kim2017} and~\cite{ben2017mutan} all propose that in the multimodal
application of VQA, the outer product of the feature
vector~$\q$ extracted from the question, with the feature vector~$\v$ extracted
from the image, produces a more expressive feature representation than
straightforward concatenation, element-wise product, or element-wise sum. These previous works design fusion operators based on approximations
to the outer product~$\q \otimes \v$.

The Multimodal Compact Bilinear Pooling (MCB) method
of~\cite{DBLP:conf/emnlp/FukuiPYRDR16} uses the Compact Bilinear Pooling
method of~\cite{gao2016compact} to approximate a bilinear interaction between
$\q$ and $\v$. In the case of MCB, the fusion operator~$\fusop$ consists of
projecting~$\q$ and~$\v$ using Count
Sketch~\cite{Charikar:2002:FFI:646255.684566}, followed by convolution of~$\q$
and~$\v$, which is done efficiently using an element-wise multiplication in
the frequency domain.

The MCB method makes use of the fact that the Count Sketch of the outer
product~$\Psi(\q \otimes \v \,, h \,, s)$, where~$h$ and~$s$ are uniform random
variables as described in~\cite{DBLP:conf/emnlp/FukuiPYRDR16}, is equal in
expectation to the convolution~$\Psi(\q \,, h \,, s) * \Psi(\v \,, h \,, s)$ of
the question and visual feature representations,
i.e.,
$E[\Psi(\q \otimes \v \,, h \,, s)] = E[\Psi(\q \,, h \,, s) * \Psi(\v \,, h \,, s)]$.
However, since this expectation is intractable to
compute,~\cite{Kim2017} proposes a fusion operator based on the Hadamard
(element-wise) product.

The fusion operator in~\cite{Kim2017}, the Multimodal Low-rank Bilinear
Attention Networks~(MLB), uses the Hadamard product coupled with projections
of~$\q$ and~$\v$ with learned weight matrices~$W_\q$ and~$W_\v$, in order to
make a low-rank approximation to the general outer product~$\q \otimes \v$. The
projected vectors are multiplied by a third weight matrix~$W_{\z}$, such that
the fusion operator in the case of~MLB
is~$\fusop{(\q \,, \v)} = {W_{\z}}^\intercal{} ({W_\q}^\intercal{} \q \odot {W_\v}^\intercal{} \v)$,
where~$\odot$ denotes the Hadamard product. Combined, the weight
matrices~$W_\q$, $W_\v$ and~$W_{\z}$ approximate the general weight tensor~$\T$
corresponding to the bilinear interaction between~$\q$ and~$\v$ in the special
case where~$\T$ is of low rank. The idea of using a constrained-rank weight
tensor in the outer product~$\q \otimes \v$ is the same idea used in the MUTAN
fusion operator of~\cite{ben2017mutan}, which generalizes MLB by allowing the
bilinear interaction tensor between~$\q$ and~$\v$ to be of rank $R$.

MUTAN is a fusion operator that is motivated
in~\cite{ben2017mutan} by the Tucker
decomposition~\cite{kolda2009tensor}, which we discuss in
Section~\ref{sec:tucker}. In Section~\ref{sec:model}, we derive the form of
bilinear interaction used by the MUTAN fusion operator from the definition of
the Tucker decomposition, since we base our fusion operator models on this
derivation as well. The MUTAN fusion operator generalizes the MLB fusion
operator to an interaction tensor of rank~$R$, and therefore is,

\begin{equation}
        \fusop{(\q \,, \v)} = \sum_{r = 1}^R {W^r_\z}^\intercal{} ({W^r_\q}^\intercal{} \q \odot {W^r_\v}^\intercal{} \v) \,.
\label{eqn:mutan-fusop}
\end{equation}

\subsection{Tucker Decomposition}
\label{sec:tucker}

\newcommand{\qtWq}{\q^\intercal{} W_\q{}}
\newcommand{\vtWv}{\v^\intercal{} W_\v{}}
\begin{figure}[!t]
\centering
\input{tikzcuboid.tikz}

\begin{tikzpicture}
        \tikzcuboid{%
            shiftx=0cm,%
            shifty=0cm,%
            scale=\cuboidscale,%
            rotation=0,%
            densityx=1,%
            densityy=1,%
            densityz=1,%
            dimx=\qvectikzdim,%
            dimy=\vvectikzdim,%
            dimz=3,%
            linefront=black,%
            linetop=black,%
            lineright=black,%
            fillfront=red!50,%
            filltop=green!50,%
            fillright=blue!50,%
            anglex=180,%
            angley=90,%
            anglez=225,%
            scalex=1,%
            scaley=0.8,%
            scalez=0.6,%
            emphedge=N,%
        }

        \tikzcuboid{%
            shiftx=2.0cm,%
            shifty=0cm,%
            scale=\cuboidscale,%
            rotation=0,%
            densityx=1,%
            densityy=1,%
            densityz=1,%
            dimx=1,%
            dimy=\vvectikzdim,%
            dimz=3,%
            linefront=black,%
            linetop=black,%
            lineright=black,%
            fillfront=red!50,%
            filltop=green!50,%
            fillright=blue!50,%
            anglex=180,%
            angley=90,%
            anglez=225,%
            scalex=1,%
            scaley=0.8,%
            scalez=0.6,%
            emphedge=N,%
        }

        \tikzcuboid{%
            shiftx=10pt,%
            shifty=-15mm,%
            scale=\cuboidscale,%
            rotation=0,%
            densityx=1,%
            densityy=1,%
            densityz=1,%
            dimx=\qvectikzdim,%
            dimy=1,%
            dimz=1,%
            linefront=black,%
            linetop=black,%
            lineright=black,%
            fillfront=red!50,%
            filltop=red!50,%
            fillright=red!50,%
            anglex=180,%
            angley=90,%
            anglez=225,%
            scalex=1,%
            scaley=0.8,%
            scalez=0.6,%
            emphedge=N,%
        }

        \newcommand{\qbracesize}{10.2ex}
        \newcommand{\vbracesize}{6.8ex}

        \coordinate (qbrace) at (-0.35, -1.77);
        \node at (qbrace) {\makebox[\qbracesize]{\upbracefill}};
        \node at (qbrace) [below] {$t_q$};

        \coordinate (Tcq brace) at (-0.42, -0.18);
        \node at (Tcq brace) {\rotatebox{180}{\makebox[\qbracesize]{\upbracefill}}};
        \node[inner sep=1pt] at (Tcq brace) [above, fill=white, fill opacity=0.8, text opacity=1, yshift=1.1mm] {$t_q$};

        \node (Tc label) at (Tcq brace) [xshift=-3mm, yshift=18mm] {$\T_c$};
        \node (Tcq label) at (Tc label) [xshift=27mm] {$\T_c^\q$};
        \node (vtWv label) at (Tc label) [xshift=45mm] {$\vtWv$};
        \node[inner sep=0pt] (qtWq label) at (Tc label) [yshift=-43mm, xshift=3mm] {$\qtWq$};
        \node (Tcqv label) at (Tc label) [xshift=65mm] {$\T_c^{\q\v}$};

        \newcommand{\Tcvy}{0.56}
        \coordinate (Tcv brace) at (-1.9, \Tcvy);
        \node at (Tcv brace) {\rotatebox{270}{\makebox[\vbracesize]{\upbracefill}}};
        \node at (Tcv brace) [left] {$t_v$};

        \coordinate (Tco brace) at (0.33, 1.05);
        \node at (Tco brace) {\rotatebox{140}{\makebox[2ex]{\upbracefill}}};
        \node at (Tco brace) [right, yshift=2.0mm, xshift=-1.0mm] {$t_o$};

        \node at (qbrace) [yshift=10mm] {$\times_1$};

        \tikzcuboid{%
            shiftx=3.8cm,%
            shifty=-7pt,%
            scale=\cuboidscale,%
            rotation=0,%
            densityx=1,%
            densityy=1,%
            densityz=1,%
            dimx=1,%
            dimy=\vvectikzdim,%
            dimz=1,%
            linefront=black,%
            linetop=black,%
            lineright=black,%
            fillfront=blue!50,%
            filltop=blue!50,%
            fillright=blue!50,%
            anglex=180,%
            angley=90,%
            anglez=225,%
            scalex=1,%
            scaley=0.8,%
            scalez=0.6,%
            emphedge=N,%
        }

        \coordinate (vbrace) at (4.1, 0.18);
        \node at (vbrace) {\rotatebox{90}{\makebox[\vbracesize]{\upbracefill}}};
        \node at (vbrace) [right] {$t_v$};

        \draw[->, thick, >= stealth] (0.85, 0.25) -- (1.35, 0.25) {};

        \node at (vbrace) [xshift=-11.5mm, yshift=0mm] {$\times_2$};

        \coordinate (tcqv pos) at (5.7, 0.31);

        \draw[->, thick, >= stealth] (4.63, 0.25) -- (5.13, 0.25) {};

        \tikzcuboid{%
            shiftx=5.76cm,%
            shifty=0.3cm,%
            scale=\cuboidscale,%
            rotation=0,%
            densityx=1,%
            densityy=1,%
            densityz=1,%
            dimx=1,%
            dimy=1,%
            dimz=3,%
            linefront=black,%
            linetop=black,%
            lineright=black,%
            fillfront=red!50,%
            filltop=green!50,%
            fillright=blue!50,%
            anglex=180,%
            angley=90,%
            anglez=225,%
            scalex=1,%
            scaley=0.8,%
            scalez=0.6,%
            emphedge=N,%
        }
\end{tikzpicture}
\caption{A geometric representation of the first two $n$-mode products of the
         right hand side Tucker decomposition of Equation~\ref{eqn:tucker-defn},
         where~$A = \qtWq$ and~$B = \vtWv$, i.e.\ $\T_c \times_1 \qtWq \times_2 \vtWv$.
         From left to right, the vector~$\qtWq$ is first combined with the core
         tensor~$\T_c$ by taking the inner product along the dimension of
         size~$t_q$ between~$\qtWq$ and each of the~$t_v \times t_o$ fibers
         composing~$\T_c$, producing the matrix $\T_c^\q$.  Then,~$\vtWv$ is
         combined with~$\T_c^\q$ by taking the inner product between~$\vtWv$
         and the~$t_o$ columns of~$\T_c^\q$, producing the
         vector~$\T_c^{\q{}\v{}}$.}
\label{fig:n-mode-prod}
\end{figure}

The Tucker decomposition is a form of higher-order principal component
analysis~\cite{kolda2009tensor}, which decomposes a tensor into a series of
three $n$-mode tensor products between a single core tensor~$\T_c$ and three
matrices:

\begin{equation}
        \T \approx \T_c \times_1 \A \times_2 \B \times_3 \C
\label{eqn:tucker-defn}
\end{equation}

In Equation~\ref{eqn:tucker-defn}, the $n$-mode tensor product denoted
by~$\times_n$ is the multiplication of a tensor by a matrix, or by a vector, as
in the case of MUTAN and our own method.

In general, the $n$-mode tensor product of a
tensor~$\T \in \R^{I_1 \times I_2 \times \cdots \times I_N}$ with a
matrix~$W \in \R^{J \times I_n}$ yields a new tensor~$\T \times_n W$ with
size~$\TsizeIn{J}$.

In our special case, where we have an $n$-mode product between~$\T$ and a
vector~$\z$, the size of~$\T \times_n \z$ becomes~$\TsizeIn{1}$. The mode-$n$
``fiber'', referring to the generalization of rows and columns of matrices to
higher-order tensors as defined in~\cite{kolda2009tensor},~$\Tfibre$ of~$\T$
along~$I_n$ becomes a single element resulting from the dot product
of~$\Tfibre$ with~$\z$.

Formally, the element-wise definition of the $n$-mode product between
tensor~$\T$ and vector~$\z$ is:

\begin{equation}
        {\big(\T \times_n \z\big)}_{i_1 \cdots i_{n - 1} i_{n + 1} \cdots i_N} = \sum_{i_n = 1}^{I_n} \T_{i_1 \cdots i_N} z_{i_n} \,,
\label{eqn:tucker-decomp-vec-z}
\end{equation}

\noindent where in Equation~\ref{eqn:tucker-decomp-vec-z}, the dimensionality
of~$\T \times_n \z$ is reduced to~$N - 1$ by summing over the~$n$th dimension.
The $n$-mode product is depicted in Figure~\ref{fig:n-mode-prod}.

In Section~\ref{sec:model}, we use the definition of the Tucker decomposition
from Equation~\ref{eqn:tucker-defn}, along with the definition of the $n$-mode
product for the special case of multiplication between a tensor and a vector,
in order to derive the MUTAN fusion operator of Equation~\ref{eqn:mutan-fusop},
and to further generalize the derived fusion operator to include Feature Gating
and Nonlinearity Ensembling.

\section{Methods}

In this section, we present the methods used to
derive a generalization of bilinear fusion operators for the combination of two
feature vectors in multimodal applications. We sub-divide our methods into the
data and model sub-categories of the machine learning process. We first discuss
the data used in our experiments, followed by discussion of the fusion operator
model and how we generalized the model into a class of fusion operators over
which we instantiated and evaluated a range of specific instances.

\subsection{Data}

The VQA~1.0 dataset~\cite{VQA} is a dataset of ``free-form and open-ended
Visual Question Answering (VQA)''. The VQA task is to give an open-ended
natural language reponse to an input consisting of a natural language question
and an image. In VQA~1.0, there are~\num{23234} unique one-word answers for
real images, and therefore the model's outputs can be represented as a multiple
choice over these answers. In practice, we limit the number of choices to the
top~\num{2000} most common answers in the training dataset.

The VQA~1.0 real images dataset consists of~\num{204}K images from the MSCOCO
dataset~\cite{kolda2009tensor},~\num{614}K questions, and~\num{6}M answers (ten
answers per question). The dataset is separated into a~$2/1/2$
training/validation/test split by the VQA~1.0 authors.

It is known that due to the data collection methodology of the VQA~1.0 dataset,
there exists an imbalance in the dataset that allows questions to be answered
with high accuracy without taking the image into
account; this is remedied with VQA~2.0~\cite{goyal2017making}, which balances the answers to each
question, so that the best scores on the VQA~2.0 dataset are not possible using
language priors alone.

The balancing of VQA~2.0 is due to the addition of complementary images, wherein
for a given image and question pair~$(I, Q)$ with answer~$A$, a complementary
image~$I'$ is found that is similar in appearance, but has a different answer~$A'$ to the
question. The new example~$(I', Q, A')$ is then added to the dataset. VQA~2.0
includes 195K, 93K, and 191K complementary images in the training, validation, and test
sets, respectively. In total, the VQA~2.0 dataset has 443K training, 214K validation, and
453K test (image, question) pairs.

The increased difficulty of the VQA~2.0 dataset with regards to emphasizing the
importance of the combination of visual and question features creates a larger
gap between relatively strong and weak fusion operators, i.e., the performance
gap between strong and weak fusion operators increases when moving from VQA~1.0
to VQA~2.0, even when the absolute performance decreases for
both models under consideration~\cite{goyal2017making}. Therefore, we evaluate
our proposed fusion operators on VQA~2.0.

\subsection{Model}
\label{sec:model}

\begin{figure*}[!t]
\centering
\input{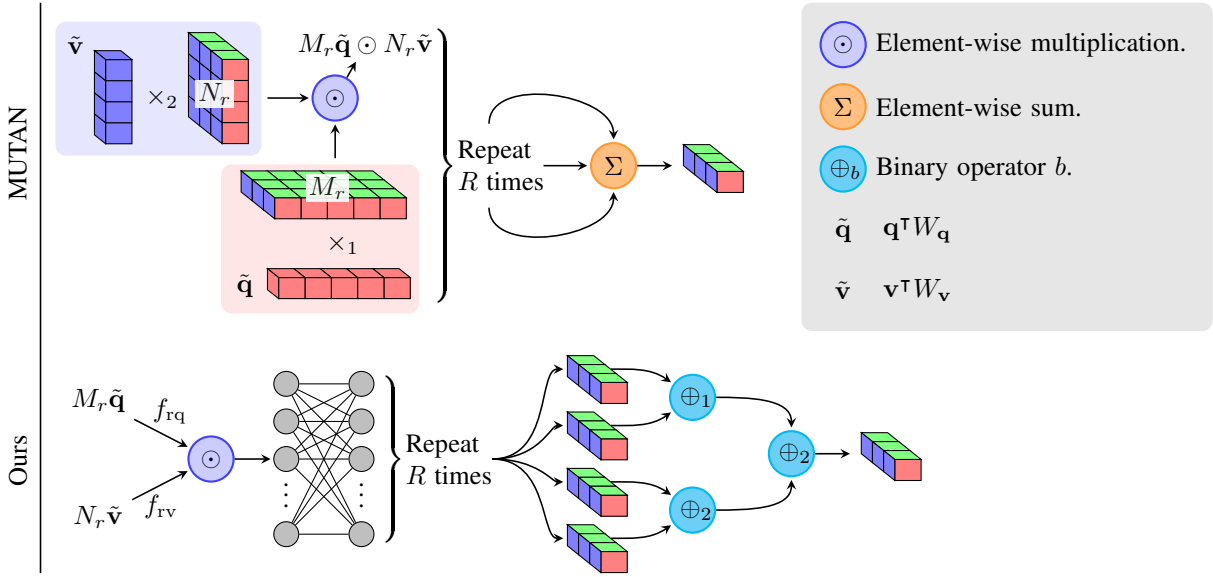}
\caption{A comparison between the MUTAN fusion operator of~\cite{ben2017mutan}
         (top) and our generalized fusion operator (bottom). Both MUTAN and our
         fusion operator take the features~$M_r\mutanfeat{q}$
         and~$N_r\mutanfeat{v}$ as input. MUTAN approximates the Tucker
         decomposition shown in Figure~\ref{fig:n-mode-prod} by constraining
         the bilinear interaction between the vectors~$\mutanfeat{q}$
         and~$\mutanfeat{v}$ to be of rank~$R$.
         Note that while~$\T_c$ in Figure~\ref{fig:n-mode-prod} is a learned
         tensor whose parameters implicitly describe the bilinear interaction
         between~$\mutanfeat{q}$ and~$\mutanfeat{v}$, here the parameters
         of~$M_r$ and~$N_r$ are separate and the bilinear interaction is due to
         the Hadamard product.
         Our fusion operator generalizes MUTAN first by allowing different
         nonlinearities~$f_{rq}$ and~$f_{rv}$ to act on the
         features~$M_r\mutanfeat{q}$ and~$N_r\mutanfeat{v}$. The result is then
         composed with an additional learned nonlinearity in the form of a
         neural network, producing a set of~$R$ output features, as in MUTAN
         after the Hadamard product
         step~$M_r\mutanfeat{q} \odot N_r\mutanfeat{v}$.  In the case of MUTAN,
         the~$R$ output feature vectors are combined by element-wise summation,
         whereas in our fusion operator the~$R$ feature vectors are combined by
         applying a tree of binary operators~$\binopb$, as described by
         Algorithm~\ref{alg:apply-binop-seq}.}
\label{fig:mutan-ours}
\end{figure*}

Following the work of~\cite{Kim2017} and~\cite{ben2017mutan}, we represent the
sentence and visual modalities with extracted feature vectors. We use a
pre-trained Skip-Thought Vectors model~\cite{DBLP:journals/corr/KirosZSZTUF15}
to extract a ``question vector'' feature representation~$\q$ from a question.
As well, we use a ResNet~\cite{he2016deep} pre-trained on
ImageNet~\cite{russakovsky2015imagenet} to extract a ``visual vector'' feature
representation~$\v$ from an image.

For the purpose of this paper, the Skip-Thought Vectors model and ResNet can
each be thought of as ``black boxes'' used to extract useful feature
representations from sentences and images, respectively. By using this
black-box abstraction, the fusion operators we develop can automatically
benefit from improved feature extraction models without altering the algorithm
presented here.

The Skip-Thought Vectors model extracts a $d_q$ vector from the question, and
the ResNet model extracts a~$d_v \times S \times S$ tensor from the image,
where~$S$ is the pre-pooling spatial dimension of the feature maps produced by
the ResNet.

The MUTAN fusion operator of~\cite{ben2017mutan} and MLB fusion operator
of~\cite{Kim2017} combine the image and sentence feature vectors by
approximating a bilinear interaction between the vectors. The bilinear
interaction is learned, and the output of that interaction is input into a
linear predictive layer, which is in turn fed into a softmax layer that outputs
probabilities over the top~\num{2000} most common answers in the training data.

We generalize the bilinear interaction of MUTAN and MLB such that both MLB and
MUTAN, as well as element-wise multiplication and element-wise addition, can
all be represented in a common form.

We derive our fusion operator as a generalization of the MUTAN fusion operator
by first deriving MUTAN from the definition of the Tucker
decomposition~\cite{kolda2009tensor}, then discussing the generalization of
this expression.  We begin from the definition of the Tucker decomposition
given in Equation~\ref{eqn:tucker-defn}.

In the case of MUTAN, $\A$ in Equation~\ref{eqn:tucker-defn} corresponds to the
vector~$\qtWq \in \R^{1 \times t_q}$, $\B$
is~$\vtWv \in \R^{1 \times t_v}$,
and~$\C$ is~$W_o \in \R^{|A| \times t_o}$, where~$|A|$ is the
dimensionality of the output vector (i.e., the number of answer classes),
and~$t_q$, $t_v$ and~$t_o$ are the respective dimensionalities of the vector
spaces that the question features, image features, and fused question-image
features are projected onto before applying a linear prediction layer~$W_o$.

Therefore, MUTAN is a special case of the Tucker decomposition in
Equation~\ref{eqn:tucker-defn}, as given by Equation~\ref{eqn:tucker-mutan},
where~$\mathbf{y}$ is the prediction output of the fusion operator:

\begin{equation}
        \mathbf{y} = \T_c \times_1
                \qtWq \times_2
                \vtWv \times_3
                W_o
        \label{eqn:tucker-mutan}
\end{equation}

In the following, we derive from the Equation~\ref{eqn:tucker-mutan} form of
MUTAN an expression for MUTAN that can be generalized to a family of fusion
operators, over which a variety of specific instantiations can be chosen and
evaluated.  Throughout the derivation, $\mutanfeat{q}$ is shorthand for
$\qtWq$, and $\mutanfeat{v}$ represents $\vtWv$.

Referring to $\T_c \times_1 \qtWq$ as $\T_c^{\q}$, each element of
the matrix $\T_c^{\q}$ in terms of elements of~$\mutanfeat{q}$ and~$\T_c$ is
$\T_c^{\q}[j, k] = \sum_{i = 1}^{t_q} \T_c[i, j, k] \cdot \mutanfeat{q}[i]$,
which follows from the definition of the $n$-mode tensor product.

Similarly, we expand the elements of the
vector~$\T_c^{\q\v} = \T_c^{\q} \mutanfeat{v}$ in terms of elements
of~$\mutanfeat{q}$ and~$\mutanfeat{v}$, and slices of~$\T_c$.

\begin{align}
\begin{split}
        \T_c^{\q\v}[k] &= \sum_{j = 1}^{t_v} \T_c^{\q}[j, k] \cdot \mutanfeat{v}[j]  \\
                       &= \sum_{j = 1}^{t_v} \biggl(\sum_{i = 1}^{t_q} \T_c[i, j, k] \cdot \mutanfeat{q}[i]\biggr) \cdot \mutanfeat{v}[j]  \\
                       &= {\mutanfeat{q}}^\intercal \T_c[:, :, k] \mutanfeat{v}
\end{split}
\label{eqn:tucker-core-tensor-expanded}
\end{align}

The constraint enforced by~\cite{ben2017mutan} in MUTAN is that each slice of
the core tensor, i.e.,~$\T_c[:, :, k]$, must be of rank~$R$, and therefore a sum
of matrices each given by an outer product of vectors. Therefore,

\begin{equation}
        \T_c[:, :, k] = \sum_{r = 1}^R \m_r^k \otimes \n_r^k
                      = \sum_{r = 1}^R \m_r^k {\n_r^k}^\intercal \,,
\label{eqn:tucker-core-tensor-rank-r}
\end{equation}

\noindent where~$\m_r^k \in \R^{t_q \times 1}$ and~$\n_r^k \in \R^{1 \times t_v}$.
Substituting Equation~\ref{eqn:tucker-core-tensor-rank-r} into the last line of
Equation~\ref{eqn:tucker-core-tensor-expanded} gives:

\begin{align}
\begin{split}
        \T_c^{\q\v}[k] &= {\mutanfeat{q}}^\intercal
                \biggl(\sum_{r = 1}^R \m_r^k {\n_r^k}^\intercal\biggr) \mutanfeat{v}  \\
                     &= \sum_{r = 1}^R \left({\mutanfeat{q}}^\intercal \m_r^k\right)
                                \left({\n_r^k}^\intercal \mutanfeat{v}\right) \,,
\end{split}
\label{eqn:tucker-core-element-wise}
\end{align}

\noindent where each term in the sum is a scalar.
Equation~\ref{eqn:tucker-core-element-wise} states that $\T_c^{\q{}\v{}}$ is the sum
over $r~\in~\{1, \dots, R\}$ of matrix-vector element-wise products
$M_r \mutanfeat{q} \odot N_r \mutanfeat{v}$, i.e.,

\begin{equation}
        \T_c^{\q{}\v{}} =
                \sum_{r = 1}^R M_r \mutanfeat{q} \odot N_r \mutanfeat{v} \,.
\label{eqn:tucker-decomp-hadamard}
\end{equation}

\noindent In Equation~\ref{eqn:tucker-decomp-hadamard}, $M_r \in \R^{t_o \times t_q}$
and~$N_r \in \R^{t_o \times t_v}$ are learned matrices, whose rows are the
vectors $\m_r^k$ and $\n_r^k$, respectively.

We have presented an equivalent form of the MUTAN fusion operator in
Equation~\ref{eqn:tucker-decomp-hadamard}, of which the MLB fusion operator
of~\cite{Kim2017} is a special case where~$R = 1$. In this paper, we propose a
further generalized extension to Equation~\ref{eqn:tucker-decomp-hadamard}, by
allowing the fusion operator to contain the following transformations:

\begin{itemize}
        \item A unique pair of unary activation
                functions~$(f_{rq}, f_{rv})$ that can wrap the individual
                factors before the Hadamard product. We refer to the
                combination of unique pairs of nonlinearities as Nonlinearity
                Ensembling.

        \item A sequence of~$L$ neural network layers~$\phi_l$ composing a
                feedforward neural network module~$\Phi$ that can take the features
                produced by the bilinear
                interaction~$M_r \mutanfeat{q} \odot N_r \mutanfeat{v}$ as
                input, thereby introducing additional nonlinearity into the
                fusion operator.

        \item Skip connections~\cite{he2016deep, srivastava2015training} may
                exist between the inputs to the fusion operator, the outputs of
                any given neural network layer~$\phi_j$ in the fusion operator,
                and the inputs to any other neural network layer~$\phi_k$, or
                the final output of the fusion operator.

        \item The sum over~$R$ terms in
                Equation~\ref{eqn:tucker-decomp-hadamard} is generalized to
                arbitrary binary relations~$\binopb(\cdot \,, \cdot)$. Each
                binary relation~$\binopb$ is associated with a
                set~$\binoppartition$ corresponding to an element of a
                partition of the outputs~$\tuckbranch$ from the~$R$ branches of
                the fusion operator, before those branches are joined.
                $\tuckbranch$ is defined according to
                Equation~\ref{eqn:general-fusion-op-defn}:

                \begin{equation}
                \tuckbranch =
                \Phi_r(f_{rq}{\left(M_r \mutanfeat{q}\right)} \odot f_{rv}{\left(N_r \mutanfeat{v}\right)}) \,.
                \label{eqn:general-fusion-op-defn}
                \end{equation}

                There is also an ordering over the binary operations such
                that~$\binopb$ form a sequence~$\binopseq$, where $B$
                represents the total number of distinct binary operators and is
                equal to the number of subsets in the
                partition~$\binoppartition$.

                The sequence~$\binopseq$ is applied recursively to the
                partitions~$\binoppartB_b \coloneqq \binoppartition$ according
                to Algorithm~\ref{alg:apply-binop-seq}.

                The generalization over binary operators allows for binary
                operations besides addition to be applied to the~$\tuckbranch$
                output from each branch of the fusion operator, and also allows
                definition of precedence rules so that binary operators can be
                applied in a defined order. The fusion operator in
                Equation~\ref{eqn:tucker-decomp-hadamard} is the special case
                of the generalized fusion operator where~$B = 1$,
                $\binop_1 = +$, and~$\binoppartB_1$ is the entire set of branch
                outputs~$\{\tuckbranch\}$.
\end{itemize}

\begin{algorithm}
\begin{algorithmic}[1]
        \Function{ApplyBinOpSequence}
                 {$\binoppartBseq, \binopseq$}
                \State{$v \gets \Call{Identity}{\binop_1}$}
                \ForAll{$b \in \{1, \dots, B\}$}
                        \State{$v_b \gets \Call{Identity}{\binopb}$}
                        \ForAll{$\tuckbranch \in \binoppartB_b$}
                                \State{$v_b \gets v_b \binopb \tuckbranch$}
                        \EndFor{}
                        \State{$v \gets v \binopb v_b$}
                \EndFor{}
                \State{\Return{$v$}}
        \EndFunction{}
\caption{Recursively applies the binary operator sequence~$\binopseq$ to the
        sequence~$\binoppartBseq$ of elements of a partition of the set of
        outputs from each of the~$R$ branches of the generalized fusion
        operator. The~$\textproc{Identity}{(\binop)}$ function returns the
        identity for the binary operator~$\binop$.}
\label{alg:apply-binop-seq}
\end{algorithmic}
\end{algorithm}


Our generalized fusion operator is compared alongside MUTAN in
Figure~\ref{fig:mutan-ours}.

\section{Experiments}
\label{sec:experiments}

In this section, we instantiate specific models based on the generalized fusion
operator described by Algorithm~\ref{alg:apply-binop-seq}. We individually
test the specific extensions for a generalized fusion operator, as described in
Section~\ref{sec:model}. We then demonstrate the degree to which the
extensions' performance gains are complementary.

In all of the following experiments, the Adam
optimizer~\cite{kingma2014adam} is used to train each model with a learning
rate of~$10^{-4}$. The models are trained for~\num{100} epochs. For the validation
set, models with the best validation accuracy are selected. For the test-dev
set, the model parameters used to generate test-dev predictions correspond to
the early-stopping epoch determined from the validation set.

Following the hyperparameter settings of~\cite{ben2017mutan}, the number of
branches used in the fusion operator is~$R = 5$. The dimensionality of the
question vector~$\q$ is the default for uni-skip Skip-Thought Vectors~$d_q =
2400$, while the visual vector~$\v$ is the~$d_v = 2048$ dimensional output of a
ResNet.  Question and visual features are projected into a~$t_q = t_v = 310$
dimensional vector space before being reprojected into a~$t_o = 510$
dimensional vector space where~$\q$ and~$\v$ are combined by Hadamard product.
The batch size used in our experiments is~\num{128}.

\subsection{Nonlinearity Ensembling (NE)}

To test the contribution of using a variety of activation functions per branch
of the fusion operator, we first conducted a grid search over possible
combinations where each~$f_{rq}$ and~$f_{rv}$ was drawn uniformly from the
following set of candidate nonlinearities: identity function, leaky
ReLU~\cite{maas_rectified_nonlinearities},
SeLU~\cite{klambauer2017self}, sigmoid, and tanh. We subsampled the VQA~1.0
dataset and ran a grid search with each run lasting only~\num{5} epochs in
order to quickly observe many combinations of nonlinear activation function
pairs.

From the grid search, we observed the pattern that the stronger nonlinearity
pair combinations used SeLU as the visual vector activation function, while
using a variety of different activation functions on the question vector.
Therefore, to test Nonlinearity Ensembling, we used branches where the visual
vectors always used the SeLU nonlinearity, and the branches used one of each
activation function on the question vector.

We propose that the diversity in the activation function on the question vector
decreases the correlation between the branches, hence improving the ensembling
effect of summing the branches' predictions together.

\subsection{Post-fusion neural networks with skip connections}

In order to allow for a nonlinear function to act on the bilinear features
extracted by the Hadamard product between~$M_r\mutanfeat{q}$
and~$N_r\mutanfeat{v}$, we propose adding a neural network~$\Phi_r$ to the
fusion operator. We embed a ``tiny'' neural network in the fusion operator
analogous to how Network in Network~\cite{lin2013network} embeds a tiny neural
network in a convolution.

In our implementation, $\Phi_r$ consists of a sequence of blocks of three
feedforward layers where the activation function for each layer matches that
of~$f_{rq}$.  There is a skip connection from the input, and from every third
layer, to the output.

\subsection{Feature Gating (FG) and Polarity Swap (PS)}

\begin{figure*}[!t]
\centering
\newcommand{\branchstub}[5]{%
\begin{scope}[yshift=#2]
\node (q#1) {$M_{#5}\mutanfeat{q}$};
\node (v#1) [below=of q#1] {$N_{#5}\mutanfeat{v}$};

\node[hadamard, xshift=2cm] (h#1) at ($(q#1)!0.5!(v#1)$) {$\odot$}
         edge [pre] node [below] {$f_{\mathrm{selu}}$} (v#1) {}
         edge [pre] node [above, xshift=#4] {$f_{\mathrm{#3}}$} (q#1) {};

\node[nn] (nn#1) [right=of h#1] {$\Phi_#5$}
        edge [pre] (h#1) {};
\end{scope}}

\newcommand{\vecelems}[1]{%
        \node [matrix, vecsty]
                (vector#1) [right=of nn#1, xshift=1cm]
                {%
                        \node[rectangle] (vec#1a) {$t_1$};\\
                        \node[rectangle] (vec#1b) {$t_2$};\\[-2mm]
                        \node[rectangle] (vec#1c) [xshift=1mm] {$\vdots$};\\
                        \node[rectangle] (vec#1d) {$t_o$};\\
                };

        \draw [post, shorten >= 3pt, red!80, thick] (nn#1) -- (vector#1);
        \fill[left color=red!80, middle color=red!65, right color=red!50] ($(nn#1)!0.5!(vector#1)$) -- (vector#1.north west) -- (vector#1.south west) -- cycle;
}

\begin{tikzpicture}
        [gate/.style={circle,
                      inner sep=0pt,
                      minimum size=2mm,
                      thick,
                      draw=black,
                      fill=white},
         gateedge/.style={-{Circle[fill=white]},
                          out=0,
                          in=180,
                          shorten >= 3pt,
                          semithick},
         hadamard/.style={circle,
                          inner sep=0pt,
                          minimum size=6mm,
                          thick,
                          draw=blue!75,
                          fill=blue!20},
         nn/.style={rectangle,
                    draw=black!50,
                    fill=black!20,
                    thick,
                    inner sep=0pt,
                    minimum size=6mm},
         pre/.style={<-, shorten <= 1pt, >=stealth, semithick},
         post/.style={->, shorten >= 1pt, >=stealth, semithick},
         sigmoid/.style={circle,
                         inner sep=0pt,
                         minimum size=4mm,
                         thick,
                         draw=black,
                         fill=green!50,
                         draw=green!80},
         sumsty/.style={circle,
                        inner sep=0pt,
                        minimum size=6mm,
                        thick,
                        draw=orange!80,
                        fill=orange!50},
         vecsty/.style={fill=red!50, draw=red!80, very thick}]

        \branchstub{1}{0cm}{tanh}{1mm}{1}
        \branchstub{2}{-2.5cm}{identity}{2mm}{\sigma}

        \branchstub{3}{-5cm}{sigmoid}{2mm}{2}

        \vecelems{1}
        \vecelems{3}

        \node[sigmoid] (sigmoid2)
                [right=of nn2, xshift=-5mm] {$\sigma$}
                edge [pre] (nn2) {}
                edge [gateedge] (vec1a) {}
                edge [gateedge] (vec1b) {}
                edge [gateedge] (vec1d) {}
                edge [gateedge] (vec3a) {}
                edge [gateedge] (vec3b) {}
                edge [gateedge] (vec3d) {};

        \node[sumsty] (sum) [right=of sigmoid2, xshift=2cm] {$\Sigma$}
              edge [pre, out=90, in=0] (vector1) {}
              edge [pre, out=270, in=0] (vector3) {};

        \node (tcqv) [right=of sum] {$\T_c^{\q\v}$}
                edge [pre] (sum) {};

        \node[matrix] (legend) at (current bounding box.east) [xshift=3.5cm] {%
                \node[label=right:Question vector features.] {$\q$};\\[2mm]
                \node[label=right:Image vector features.] {$\v$};\\[2mm]
                \node[hadamard, label=right:Element-wise multiplication.] {$\odot$};\\[2mm]
                \node[nn, label=right:Neural network layers~${(\phi_l)}_r$.] {$\Phi_r$};\\[2mm]
                \node[sigmoid, label=right:Logistic sigmoid nonlinearity.] {$\sigma$};\\[2mm]
                \node[gate, label=right:Scalar multiplication.] {};\\[2mm]
                \node[vecsty, label=right: Elements of~$\tuckbranch \in \mathbb{R}^{t_o}$.] {$t_i$};\\[2mm]
                \node[sumsty, label=right: Element-wise sum.] {$\Sigma$};\\
        };

        \begin{scope}[on background layer]
                \node (r1) [fill=black!10, rounded corners, fit=(legend)] {};
        \end{scope}
\end{tikzpicture}
\caption{An example network using the Feature Gating neural
         network component, where each node represents a computation and
         the arrows represent the forward flow of information. The question and
         image feature vectors~$\q$ and~$\v$ are shared inputs to all branches.
         The~$\Phi_r$ nodes represent post-fusion feedforward neural networks
         with skip connections. The logistic sigmoid node~$\sigma$ squashes
         output features~$\T_\sigma^{\q{}\v{}}$ from~$\Phi_\sigma$ to a vector
         of values in~$(0, 1)$. The output from~$\sigma$ is element-wise
         multiplied with all other~$\tuckbranch$ features from each branch,
         effectively turning on or off each feature channel. The resultant
         gated~$\tuckbranch$ features are summed to become~$\T_c^{\q{}\v{}}$,
         features that are input into a predictive layer to score the most
         common answers to questions from the VQA task.}
\label{fig:feature-gating}
\end{figure*}
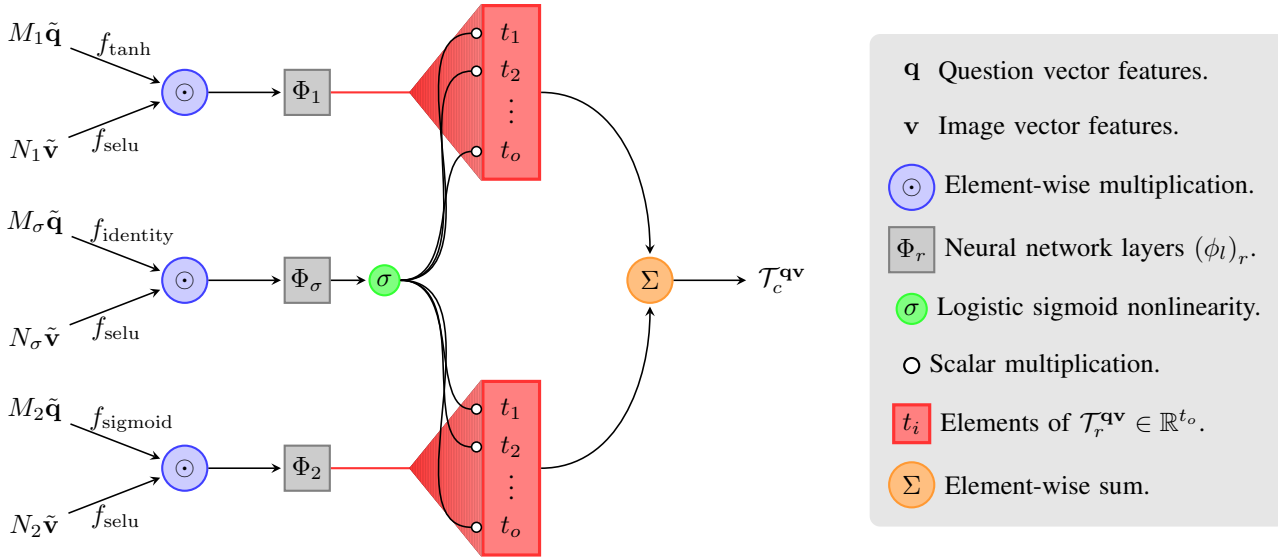

Algorithm~\ref{alg:apply-binop-seq} introduces a method of generalizing the sum
operation over fusion operator branch outputs~$\tuckbranch$ to an ordered set
of binary operations. We test two particular instantiations of
Algorithm~\ref{alg:apply-binop-seq} where~$B = 2$, $\binop_1 = +$,
$\binop_2 = \odot$ and $|\binoppartB_2| = 1$, i.e.~$R - 1$ of the branch
outputs~$\tuckbranch$ are summed and then element-wise multiplied by the
remaining branch output.

Assume that the branch output that gets multiplied with the sum of the other
branch outputs is~$\T_R^{\q{}\v{}}$. The output of~$\T_R^{\q{}\v{}}$ is first
squashed by a nonlinearity~$f$ before being multiplied into the sum
over~$\binoppartB_1$.

In the case of our first experiment, the squashing nonlinearity is the logistic
function~$f_\mathrm{sigmoid}$, which independently squashes the
output~$f_\mathrm{sigmoid}(\T_R^{\q{}\v{}})$ into the range~$(0, 1)$. The
effect of multiplying by this squashed output is therefore to turn each feature
on or off, so we refer to the first experiment as Feature Gating.

The squashing nonlinearity used in the second experiment is
tanh~$f_{\tanh}$, which has the effect of
squashing~$f_{\tanh}(\T_R^{\q{}\v{}})$ to the range~$(-1, 1)$. We refer to the
second experiment as Polarity Swap, since negative features can be
conditionally swapped to positive and vice-versa.

Figure~\ref{fig:feature-gating} demonstrates an example instantiated fusion
operator that makes use of the Feature Gating idea, and implements the
Feature Gating experiment described above. In Figure~\ref{fig:feature-gating},
the number of branches is set to~$R = 3$ for clarity, while in the experiments
of Table~\ref{tab:model-ablation}, the models have~$R = 5$ branches.

\section{Results}

In Table~\ref{tab:model-ablation}, we compare performance improvements between
different instantiations of the generalized fusion operator described in
Equation~\ref{eqn:general-fusion-op-defn} and
Algorithm~\ref{alg:apply-binop-seq}, which are discussed in
Section~\ref{sec:experiments}. We use the VQA~1.0 validation set to compare the
effect on the performance of fusion operators of adding Nonlinearity
Ensembling, as well as Feature Gating and Polarity Swap, independently. We
chose to first investigate these static components of the fusion operator
design before fixing them while investigating possible post-fusion neural
networks.  Since the neural network is a learned component of the architecture,
it has to adapt to the static components during training, and hence the optimal
hyperparameters of the neural network may vary depending on the choice of
static components in the model.

\begin{table}[!t]
\centering
\caption{An ablation study on Nonlinearity Ensembling, Feature Gating, and
         Polarity Swap.}
\begin{tabular}{rc}
\textbf{Model} & \textbf{VQA 1.0 val} \\
\midrule
MUTAN~\cite{ben2017mutan} & 61.54 \\
\midrule
Nonlinearity Ensembling (NE) & 61.66 \\
Feature Gating (FG) & 61.72 \\
NE + Polarity Swap (PS) & 61.77 \\
\textbf{NE + FG} & \textbf{61.86} \\
\end{tabular}
\label{tab:model-ablation}
\end{table}

The Nonlinearity Ensembling component improves the fusion operator's
performance on the VQA~1.0 validation set, and this improvement stacks with
the performance improvement achieved from Feature Gating. Furthermore, the
performance of Feature Gating is better compared to Polarity Swap, when each
is combined with Nonlinearity Ensembling.

We find that the improvements from the Feature Gating and Polarity Swap model
elements do not stack. We combine the ideas by using one branch each to
multiply by a vector of tanh and sigmoid outputs. The combined model performs
slightly worse than the Polarity Swap model by itself, which in turn performs
worse than the Feature Gating model.

Therefore, the best performing combination amongst Nonlinearity Ensembling,
Feature Gating and Polarity Swap occurs from using the Nonlinearity Ensembling
and Feature Gating together.

Building on the best model based on the ablation study of
Table~\ref{tab:model-ablation}, we use the more challenging VQA~2.0 validation
set to evaluate different post-fusion neural network architectures. We find
that a post-fusion neural network with six layers and~\num{128} hidden units
per layer outperforms the NE + FG model by a margin of~$0.53$ percentage
points, improving the VQA~2.0 validation OpenEnded accuracy (as defined
in~\cite{goyal2017making}) from~$60.57\%$ to~$61.1\%$. We find that with the
low number of~\num{128} hidden units, dropout is detrimental to the accuracy,
and the best model does not use dropout.

In Table~\ref{tab:sota-comparison}, we evaluate our best model on the VQA~2.0
test-dev and test-std sets, and compare to previous state of the art models
upon which our work is based, as well as to the best models
of~\cite{teney2017tips}, the winners of the 2017 VQA challenge. We
find that our best model achieves an absolute percentage point improvement
of~$1.1\%$ over the strong baseline of~\cite{ben2017mutan}. We note that the
improvement in OpenEnded accuracy of our model on the test-dev set is
significantly larger in magnitude when compared to the improvement
of~\cite{ben2017mutan} over~\cite{DBLP:conf/emnlp/FukuiPYRDR16}. We attribute
our relatively large improvement in accuracy to the introduction of
nonlinearities in the fusion operator. The nonlinearities both allow the fusion
operator to ensemble nonlinear functions of the input features (via
Nonlinearity Ensembling), and to model nonlinear relationships between the
bilinear features extracted by the Hadamard product (via the post-fusion neural
networks).

When comparing our model's performance to that of the models
of~\cite{teney2017tips} in Table~\ref{tab:sota-comparison}, we note that the
best performing models of~\cite{teney2017tips} gain a significant performance
boost from using superior image features for the VQA task. In
particular,~\cite{teney2017tips} use bottom-up
attention~\cite{anderson2017bottom}, which makes use of a Faster
R-CNN~\cite{ren2015faster} pipeline, to obtain features from object proposal
regions of an image. Bottom-up attention features improve the models
of~\cite{teney2017tips} by~$\approx 3\%$ absolute percentage points on average,
and since our contribution focuses solely on improving the fusion operator, our
model should gain similar improvements from using bottom-up attention features.
The multimodal fusion operator used in~\cite{teney2017tips} is a Hadamard
product, as in MLB\@.

\begin{table*}[!t]
\centering
\caption{A comparison with the state of the art of our best single model on the
         VQA~2.0 test-dev and test-std sets.}
\begin{tabular}{l*{9}{c}}
& \multicolumn{4}{c}{\textbf{VQA~2.0 test-dev}} & & \multicolumn{4}{c}{\textbf{VQA~2.0 test-std}} \\
\textbf{Model} & All & Y/N & Number & Other & & All & Y/N & Number & Other \\
\midrule
MCB~\cite{DBLP:conf/emnlp/FukuiPYRDR16} as reported in~\cite{goyal2017making} & 61.96 & 78.41 & 38.81 & 53.23 & & 62.27 & 78.82 & 38.28 & 53.36 \\
MUTAN~\cite{ben2017mutan} as trained and evaluated by us & 63.13 & 80.7 & 39.4 & 53.55 & & --- & --- & --- & --- \\
ResNet features~$7 \times 7$ (single)~\cite{teney2017tips} & 62.07 & 79.20 & 39.46 & 52.62 & & 62.27 & 79.32 & 39.77 & 52.59 \\
Bottom-up attention image features, adaptive~$K$ (single)~\cite{teney2017tips} & 65.32 & 81.82 & 44.21 & 56.05 & & 65.67 & 82.20 & 43.90 & 56.26 \\
\midrule
Ours (single) & 64.22 & 81.19 & 40.95 & 55.05 & & 64.64 & 81.62 & 41.19 & 55.22 \\
\midrule
\end{tabular}
\label{tab:sota-comparison}
\end{table*}

\section{Conclusion and Future Work}

We presented a generalization of the MUTAN operator aimed at fusing multi-modal
representations. We demonstrated the expressibility of the operator, showing
that it could ensemble a variety of different nonlinearities, implement neural
networks post-ensembling, and generalize MUTAN's sum-based reduction to
hierarchical fusion with arbitrary binary operators. The few configurations we
tested demonstrated significant gains relative to MUTAN and MCB on the VQA 2.0
task. However, the most promise lies not in choosing a configuration by hand,
but leveraging this generalization as a design space for architecture search by
reinforcement learning or other methods. This is the focus of future work.

\FloatBarrier{}




\bibliographystyle{IEEEtran}
\bibliography{variations_on_vqa_hadamard_product}

\begin{thebibliography}{10}
\providecommand{\url}[1]{#1}
\csname url@samestyle\endcsname
\providecommand{\newblock}{\relax}
\providecommand{\bibinfo}[2]{#2}
\providecommand{\BIBentrySTDinterwordspacing}{\spaceskip=0pt\relax}
\providecommand{\BIBentryALTinterwordstretchfactor}{4}
\providecommand{\BIBentryALTinterwordspacing}{\spaceskip=\fontdimen2\font plus
\BIBentryALTinterwordstretchfactor\fontdimen3\font minus
  \fontdimen4\font\relax}
\providecommand{\BIBforeignlanguage}[2]{{%
\expandafter\ifx\csname l@#1\endcsname\relax
\typeout{** WARNING: IEEEtran.bst: No hyphenation pattern has been}%
\typeout{** loaded for the language `#1'. Using the pattern for}%
\typeout{** the default language instead.}%
\else
\language=\csname l@#1\endcsname
\fi
#2}}
\providecommand{\BIBdecl}{\relax}
\BIBdecl

\bibitem{zoph2016neural}
B.~Zoph and Q.~V. Le, ``Neural architecture search with reinforcement
  learning,'' in \emph{5th Int. Conf. on Learning Representations (ICLR)},
  2017.

\bibitem{bello2017neural}
I.~Bello, B.~Zoph, V.~Vasudevan, and Q.~V. Le, ``Neural optimizer search with
  reinforcement learning,'' in \emph{Int. Conf. on Machine Learning (ICML)},
  2017.

\bibitem{ramachandran2017searching}
P.~Ramachandran, B.~Zoph, and Q.~Le, ``Searching for activation functions,''
  \emph{arXiv preprint arXiv:1710.05941}, 2017.

\bibitem{liu2018hierarchical}
H.~Liu, K.~Simonyan, O.~Vinyals, C.~Fernando, and K.~Kavukcuoglu,
  ``Hierarchical representations for efficient architecture search,''
  \emph{Int. Conf. on Learning Representations (ICLR)}, 2018.

\bibitem{malkomes2016bayesian}
G.~Malkomes, C.~Schaff, and R.~Garnett, ``Bayesian optimization for automated
  model selection,'' in \emph{Advances in Neural Information Processing Systems
  29 (NIPS)}, 2016, pp. 2900--2908.

\bibitem{DBLP:conf/emnlp/FukuiPYRDR16}
A.~Fukui, D.~H. Park, D.~Yang, A.~Rohrbach, T.~Darrell, and M.~Rohrbach,
  ``Multimodal compact bilinear pooling for visual question answering and
  visual grounding,'' in \emph{Proc. 2016 Conf. on Empirical Methods in Natural
  Language Processing (EMNLP)}, 2016, pp. 457--468.

\bibitem{Kim2017}
J.-H. Kim, K.~W. On, W.~Lim, J.~Kim, J.-W. Ha, and B.-T. Zhang, ``{Hadamard
  Product for Low-rank Bilinear Pooling},'' in \emph{The 5th Int. Conf. on
  Learning Representations (ICLR)}, 2017.

\bibitem{ben2017mutan}
H.~Ben-younes, R.~Cadene, M.~Cord, and N.~Thome, ``{MUTAN}: Multimodal tucker
  fusion for visual question answering,'' in \emph{IEEE Int. Conf. on Computer
  Vision (ICCV)}, 2017.

\bibitem{he2016deep}
K.~He, X.~Zhang, S.~Ren, and J.~Sun, ``Deep residual learning for image
  recognition,'' in \emph{Proc. IEEE Conf. on Computer Vision and Pattern
  Recognition (CVPR)}, 2016, pp. 770--778.

\bibitem{DBLP:journals/corr/KirosZSZTUF15}
R.~Kiros, Y.~Zhu, R.~Salakhutdinov, R.~S. Zemel, A.~Torralba, R.~Urtasun, and
  S.~Fidler, ``Skip-thought vectors,'' in \emph{Advances in Neural Information
  Processing Systems 28 (NIPS)}, 2015, pp. 3294--3302.

\bibitem{gao2016compact}
Y.~Gao, O.~Beijbom, N.~Zhang, and T.~Darrell, ``Compact bilinear pooling,'' in
  \emph{Proc. IEEE Conf. on Computer Vision and Pattern Recognition (CVPR)},
  2016, pp. 317--326.

\bibitem{Charikar:2002:FFI:646255.684566}
M.~Charikar, K.~Chen, and M.~Farach-Colton, ``Finding frequent items in data
  streams,'' in \emph{Proc. 29th Int. Colloquium on Automata, Languages and
  Programming}, 2002, pp. 693--703.

\bibitem{kolda2009tensor}
T.~G. Kolda and B.~W. Bader, ``Tensor decompositions and applications,'' in
  \emph{SIAM review}, 2009, pp. 455--500.

\bibitem{VQA}
S.~Antol, A.~Agrawal, J.~Lu, M.~Mitchell, D.~Batra, C.~L. Zitnick, and
  D.~Parikh, ``{VQA}: {V}isual {Q}uestion {A}nswering,'' in \emph{Int. Conf. on
  Computer Vision (ICCV)}, 2015.

\bibitem{goyal2017making}
Y.~Goyal, T.~Khot, D.~Summers-Stay, D.~Batra, and D.~Parikh, ``Making the v in
  vqa matter: Elevating the role of image understanding in visual question
  answering,'' in \emph{Proc. IEEE Conf. on Computer Vision and Pattern
  Recognition (CVPR)}, 2017.

\bibitem{russakovsky2015imagenet}
O.~Russakovsky, J.~Deng, H.~Su, J.~Krause, S.~Satheesh, S.~Ma, Z.~Huang,
  A.~Karpathy, A.~Khosla, M.~Bernstein \emph{et~al.}, ``Imagenet large scale
  visual recognition challenge,'' in \emph{Int. Journal of Computer Vision},
  2015, pp. 211--252.

\bibitem{srivastava2015training}
R.~K. Srivastava, K.~Greff, and J.~Schmidhuber, ``Training very deep
  networks,'' in \emph{Advances in Neural Information Processing Systems 28
  (NIPS)}, 2015, pp. 2377--2385.

\bibitem{kingma2014adam}
D.~P. Kingma and J.~Ba, ``Adam: A method for stochastic optimization,'' in
  \emph{3rd Int. Conf. on Learning Representations (ICLR)}, 2015.

\bibitem{maas_rectified_nonlinearities}
A.~L. Maas, A.~Y. Hannun, and A.~Y. Ng, ``Rectifier nonlinearities improve
  neural network acoustic models,'' in \emph{Int. Conf. on Machine Learning
  (ICML)}, 2013.

\bibitem{klambauer2017self}
G.~Klambauer, T.~Unterthiner, A.~Mayr, and S.~Hochreiter, ``Self-normalizing
  neural networks,'' in \emph{Advances in Neural Information Processing Systems
  (NIPS)}, 2017, pp. 972--981.

\bibitem{lin2013network}
M.~Lin, Q.~Chen, and S.~Yan, ``Network in network,'' \emph{Int. Conf. on
  Learning Representations (ICLR)}, 2014.

\bibitem{teney2017tips}
D.~Teney, P.~Anderson, X.~He, and A.~v.~d. Hengel, ``Tips and tricks for visual
  question answering: Learnings from the 2017 challenge,'' \emph{arXiv preprint
  arXiv:1708.02711}, 2017.

\bibitem{anderson2017bottom}
P.~Anderson, X.~He, C.~Buehler, D.~Teney, M.~Johnson, S.~Gould, and L.~Zhang,
  ``Bottom-up and top-down attention for image captioning and vqa,''
  \emph{arXiv preprint arXiv:1707.07998}, 2017.

\bibitem{ren2015faster}
S.~Ren, K.~He, R.~Girshick, and J.~Sun, ``Faster r-cnn: Towards real-time
  object detection with region proposal networks,'' in \emph{Advances in Neural
  Information Processing Systems (NIPS)}, 2015, pp. 91--99.

\end{thebibliography}

\end{document}